\documentclass[conference]{IEEEtran}
\IEEEoverridecommandlockouts
\usepackage{cite}
\usepackage{amsmath,amssymb,amsfonts}
\usepackage{algorithmic}
\usepackage{graphicx}
\usepackage{xcolor}
\usepackage{float}
\usepackage{makecell}
\def\BibTeX{{\rm B\kern-.05em{\sc i\kern-.025em b}\kern-.08em
    T\kern-.1667em\lower.7ex\hbox{E}\kern-.125emX}}
\begin{document}

\title{Stabilized Adaptive Loss and Residual-Based Collocation for Physics-Informed Neural Networks\\

}

\author{
\IEEEauthorblockN{
Divyavardhan Singh, 
Shubham Kamble, 
Dimple Sonone, 
Kishor Upla
}
\IEEEauthorblockA{
\textit{Sardar Vallabhbhai National Institute of Technology (SVNIT), Surat, India} \\
\{divyavardhansingh2004, 
shubhamkamble200431, 
dimple2423sonone, 
kishorupla\}@gmail.com
}
}
\maketitle

\begin{abstract}
Physics-Informed Neural Networks (PINNs) have been recognized as a mesh-free
alternative to solve partial differential equations where physics information
is incorporated~\cite{raissi2019,raissi2020}. However, in dealing with
problems characterized by high stiffness or shock-dominated dynamics,
traditional PINNs have been found to have limitations, including unbalanced
training and inaccuracy in solution, even with small physics
residuals~\cite{wang2021,shin2020,mishra2022}. In this research, we seek to
address these limitations using the viscous Burgers' equation with low
viscosity~\cite{whitham1974,gottlieb1997} and the Allen-Cahn
equation~\cite{allen1979} as test problems. In addressing unbalanced training,
we have developed a new adaptive loss balancing scheme using smoothed gradient
norms to ensure satisfaction of initial and boundary
conditions~\cite{xiang2022,jagtap2020}. Further, to address inaccuracy in
the solution, we have developed an adaptive residual-based collocation scheme
to improve the accuracy of solutions in the regions with high physics
residuals~\cite{wu2023,yu2022}. The proposed new approach significantly
improves solution accuracy with consistent satisfaction of physics residuals.
For instance, in the case of Burgers' equation, the relative $L_2$ error is
reduced by about 44\% compared to traditional PINNs, while for the
Allen-Cahn equation, the relative $L_2$ error is reduced by approximately
70\%. Additionally, we show the trustworthy solution comparison of the
proposed method using a robust finite difference solver~\cite{leveque2007}.
\end{abstract}


\section{Introduction}

Physics-Informed Neural Networks (PINNs), first introduced by Raissi et
al.~\cite{raissi2019}, have evolved into a powerful tool for tackling partial
differential equations by integrating the laws of physics into the training of
neural networks~\cite{raissi2020}. This is done using automatic
differentiation, which incorporates the fundamental equations, initial
conditions, and boundary conditions within a single optimization step. This
mesh-free technique has provided a considerable amount of flexibility compared
to other numerical methods~\cite{weinan2018,sirignano2018}, opening up new
avenues for a wide range of scientific computing problems, such as fluid
dynamics, reaction-diffusion problems, inverse problems, and data-informed
discovery of physical laws~\cite{raissi2020,han2018}. Despite the success of
PINNs~\cite{jagtap2020,lu2021}, it has also been observed that this technique
struggles with stiff problems involving shock-dominated
PDEs~\cite{wang2021,shin2020,mishra2022}. This is because, in such problems,
the underlying equations exhibit large gradients, multiple scales, and low
viscosity~\cite{whitham1974,gottlieb1997}, which makes the optimization
landscape difficult to handle. A common phenomenon that has puzzled many
researchers using PINNs is that, although the physics-informed residuals
converge to a very small value, the solution can be globally
incorrect~\cite{wang2021,shin2020,mishra2022}. This has also raised many
questions about the reliability of PINNs for tackling difficult nonlinear
PDEs. Stiff PDEs are characterized by rapid variations in the solution and
strong sensitivity to perturbations. Canonical examples include the viscous
Burgers' equation in the low-viscosity regime~\cite{whitham1974,gottlieb1997},
\begin{equation}
\frac{\partial u}{\partial t} + u \frac{\partial u}{\partial x} =
\nu \frac{\partial^2 u}{\partial x^2},
\end{equation}
and the Allen-Cahn equation~\cite{allen1979},
\begin{equation}
\frac{\partial u}{\partial t} - \varepsilon^2 \frac{\partial^2 u}{\partial x^2}
+ u^3 - u = 0,
\end{equation}
both of which develop sharp solution features that are difficult to capture
using smooth neural representations. In these settings, standard PINNs
frequently exhibit training instability, weak enforcement of boundary
conditions, and excessive smoothing of physically meaningful shocks or
interfaces~\cite{wang2021,gottlieb1997}.

An important underlying idea of this work is that having small PDE residuals
does not necessarily mean that the solution is correct~\cite{mishra2022}. In
fact, for multi-term PINN loss functions, the PDE residual, initial condition
loss, and boundary condition loss can be on vastly different scales, applying
gradients of vastly different magnitudes. This can lead to unbalanced
optimization, where some loss terms dominate while others are
ignored~\cite{wang2021}. This can cause PINNs to converge to the solution in
a residual sense, but not on the correct solution manifold. To address the
issues with the basic PINN formulation, a number of different modifications
have been suggested. For instance, an adaptive loss weighting can be used to
balance gradients~\cite{xiang2022,jagtap2020}, while adaptive sampling and
collocation can be used to refine the solution where errors are
large~\cite{wu2023,yu2022}. Although promising, these methods have typically
only been studied individually. 
Importantly, this work argues that the basic problem with PINNs for stiff
PDEs is not only a single problem; however, it incorporates a combination of
problems, namely that the optimization is unbalanced while, at the same time,
localized solution details are not being resolved
adequately~\cite{wang2021,mishra2022}. 

Thus, we propose a single, physics-informed learning framework that combines
the benefits of stabilized, adaptive loss balancing with those of
residual-based, adaptive collocation point
placement~\cite{xiang2022,wu2023}. The adaptive loss balancing method
dynamically adjusts the influence of the PDE residual, as well as initial and
boundary conditions, during the learning process, ensuring that the system
does not fall prey to weight collapse via gradient smoothing and
normalization~\cite{jagtap2020,wang2021}. Further, the residual-based
collocation point placement strategy, on the other hand, helps the system
focus more intently on regions of high physical
error~\cite{wu2023,yu2022}, allowing the neural network to better capture
the gradients and shock-like features of the
solution~\cite{gottlieb1997}. Together, these components drive the system
toward satisfying the PDEs and converging toward the correct solution. In the
proposed framework, we evaluate using two stiff PDE problems: the
low-viscosity viscous Burgers equation~\cite{whitham1974}, as well as the
Allen--Cahn equation~\cite{allen1979}. To ensure that the proposed evaluation
is both thorough and reliable, we use finite-difference-based PDE solvers as
references~\cite{leveque2007}, validating our results against these
established, stable references to help us distinguish between actual accuracy
improvements and those that are merely achieved by reducing the residual
error. Thus, by jointly addressing optimization dynamics and sampling
efficiency, the proposed approach advances the robustness and practical
applicability of physics-informed neural networks in challenging nonlinear
regimes~\cite{raissi2019,lu2021,wu2023}. Therefore, the main contributions
of this work are summarized as follows:
\begin{itemize}
    \item The analysis of the fundamental disconnection between physics
    residual minimization and solution accuracy in PINNs applied to stiff
    nonlinear PDEs~\cite{shin2020,mishra2022}.
    \item The proposed stabilized gradient-based adaptive loss balancing
    strategy improves training stability and constraint
    enforcement~\cite{xiang2022,jagtap2020}.
    \item Additionally, we integrate adaptive loss weighting with
    residual-based adaptive collocation within a unified framework to address
    both optimization imbalance and resolution
    limitations~\cite{wu2023,yu2022}.
    \item Finally, we showcase the significant accuracy improvements on the
    viscous Burgers' and Allen-Cahn equations, validated against stable
    numerical reference solutions~\cite{whitham1974,allen1979,leveque2007}.
    The results highlight that a reliable solution of stiff PDEs with PINNs
    requires more than physics enforcement
    alone~\cite{wang2021,mishra2022}.
\end{itemize}
\vspace{-0.3cm}

\section{Related Work}
\vspace{-0.2cm}

Prior work on improving PINNs for stiff PDEs has proceeded along two largely
independent lines: adaptive loss weighting to stabilize optimization and
adaptive collocation to improve resolution. We review each in turn and
identify the gap that motivates the present work.

\subsection{Adaptive Loss Weighting and Optimization in PINNs}

A major challenge with training PINNs is the multi-term loss function that
combines PDE residuals with initial and boundary condition constraints.
The different terms have different magnitudes and gradient strengths, which
allows some terms to dominate over others, degrading training stability and
accuracy~\cite{wang2021,shin2020}.
Wang et al.~\cite{wang2021} identified gradient flow pathologies in PINNs,
showing that imbalanced gradients cause certain loss components to dominate,
suppressing the enforcement of boundary and initial conditions.
Subsequent work in~\cite{xiang2022,jagtap2020} introduced adaptive weighting
strategies that rescale individual loss terms during training, using gradient
magnitudes, loss ratios, or heuristic balancing rules. Furthermore, the
multi-task learning and gradient normalization techniques have also been
explored, improving optimization stability and boundary
satisfaction~\cite{wang2021}, but introducing a new failure mode: loss-weight
collapse, in which a single loss term drives the optimizer while the remaining
terms are ignored~\cite{wang2021,mishra2022}.
Critically, even when adaptive weighting prevents collapse, improved
optimization stability does not reliably translate to improved solution
accuracy~\cite{mishra2022,shin2020}, particularly for stiff PDEs with
shock-like gradients~\cite{whitham1974,gottlieb1997}.
This indicates that optimization-centered approaches alone are insufficient
for stiff problems~\cite{wang2021}.
\vspace{-0.2cm}

\subsection{Adaptive Sampling and Collocation Strategies}
\vspace{-0.2cm}

A complementary direction for improving PINN performance is to adaptively
place collocation points where the PDE residual is
large~\cite{wu2023,yu2022}. The original PINN formulation sampled
collocation points uniformly at random, treating all regions of the domain
with equal importance~\cite{raissi2019}. Subsequent work demonstrated that
the spatial distribution of collocation points significantly affects solution
quality, especially near steep gradients and shock
fronts~\cite{wu2023,gottlieb1997,whitham1974}. Residual-based adaptive refinement (RAR) and residual-based adaptive
distribution (RAD) concentrate points in high-residual regions, improving
resolution where the network struggles most~\cite{wu2023}.
More recent frameworks such as PINNACLE extend this idea with systematic
criteria for point placement near sharp interfaces and complex
dynamics~\cite{wu2023,yu2022}.
These strategies yield meaningful gains in problems with localized errors,
but they do not address the underlying multi-term optimization imbalance
in the PINN loss~\cite{wang2021,mishra2022}.
As a result, adaptive collocation applied in isolation may refine the solution
in high-residual regions while the global optimization remains
unstable~\cite{wang2021}.

The two lines of research reviewed above are typically studied in isolation
and evaluated against reference solutions that are sometimes inaccurate
themselves~\cite{mishra2022,shin2020}. Thus, there is a scarcity of work
that has systematically combined stabilized adaptive loss balancing with
residual-based adaptive collocation and validated the combined approach
against a numerically reliable reference on stiff nonlinear PDEs. This paper
addresses that gap. The proposed work applies a stabilized, gradient-norm-based
adaptive loss weighting scheme together with residual-based adaptive
collocation~\cite{xiang2022,jagtap2020,wu2023}, and rigorously evaluates
the combined effect on solution accuracy using a stable finite-difference
reference for the Burgers' equation and boundary-error metrics for the
Allen-Cahn equation~\cite{whitham1974,leveque2007}.
The central finding of the proposed work is that neither technique alone is
sufficient: robust accuracy on stiff PDEs requires both optimization
stabilization and resolution-aware sampling working in tandem.
\vspace{-0.2cm}
\section{Problem Formulation}
\vspace{-0.2cm}

The Physics-Informed Neural Networks (PINNs) solve the nonlinear
time-varying partial differential equations defined on a bounded domain
in space and time~\cite{whitham1974} by employing a neural network to
approximate the unknown field and a combined loss function to enforce the
PDE along with initial and boundary
conditions~\cite{raissi2019,raissi2020,cybenko1989}.

\vspace{-0.2cm}
\subsection{Governing Equations}

\subsubsection{Viscous Burgers' Equation}

The one-dimensional viscous Burgers' equation~\cite{whitham1974} is given by
\vspace{-0.1cm}
\begin{equation}
\frac{\partial u(x,t)}{\partial t}
+ u(x,t)\frac{\partial u(x,t)}{\partial x}
- \nu \frac{\partial^2 u(x,t)}{\partial x^2}
= 0,
\label{eq:burgers}
\end{equation}
\vspace{-0.1cm}
where $u(x,t)$ denotes the solution field and $\nu > 0$ is the viscosity
coefficient. In this work, a low-viscosity regime $\nu = 0.01$ is
considered, which leads to stiff dynamics and sharp solution
gradients~\cite{gottlieb1997}.
Additionally, the spatial and temporal domains are defined as
\vspace{-0.1cm}
\begin{equation}
x \in [-1,\,1], \quad t \in [0,\,1].
\end{equation}
\vspace{-0.1cm}
The initial condition is prescribed as
\vspace{-0.1cm}
\begin{equation}
u(x,0) = -\sin(\pi x),
\label{eq:burgers_ic}
\end{equation}
\vspace{-0.1cm}
homogeneous. Dirichlet boundary conditions are imposed:
\vspace{-0.1cm}
\begin{equation}
u(-1,t) = 0, \quad u(1,t) = 0.
\label{eq:burgers_bc}
\end{equation}

\subsubsection{Allen--Cahn Equation}

The Allen-Cahn equation is a nonlinear reaction-diffusion
equation~\cite{allen1979} given by
\vspace{-0.1cm}
\begin{equation}
\frac{\partial u(x,t)}{\partial t}
- \varepsilon^2 \frac{\partial^2 u(x,t)}{\partial x^2}
+ u(x,t)^3 - u(x,t) = 0,
\label{eq:allen_cahn}
\end{equation}
\vspace{-0.1cm}
where $\varepsilon$ controls the interface thickness. This equation
exhibits metastable behavior and sharp transition layers, making it a
challenging benchmark for PINNs~\cite{allen1979,raissi2019}.
The spatial and temporal domains are defined as
\vspace{-0.1cm}
\begin{equation}
x \in [-1,\,1], \quad t \in [0,\,1].
\end{equation}
\vspace{-0.1cm}
The initial and boundary conditions for the Allen-Cahn equation are
chosen following standard benchmark settings commonly used in prior PINN
studies to ensure well-posedness and fair
comparison~\cite{raissi2019,lu2021,allen1979}.

\subsection{Physics-Informed Neural Network Approximation}

A PINN approximates the solution $u(x,t)$ using a neural network
$u_{\theta}(x,t)$ with parameters $\theta$~\cite{raissi2019,cybenko1989}.
Automatic differentiation is employed to compute the required spatial and
temporal derivatives~\cite{raissi2019,lu2021}.

The physics residual for a generic PDE operator $\mathcal{N}(\cdot)$ is
defined as follows:
\vspace{-0.1cm}
\begin{equation}
f_{\theta}(x,t) = \mathcal{N}\big(u_{\theta}(x,t)\big).
\end{equation}
\vspace{-0.1cm}
For the viscous Burgers' equation, the residual explicitly takes the
form~\cite{raissi2019,whitham1974}
\vspace{-0.1cm}
\begin{equation}
f_{\theta}^{\mathrm{B}} =
\frac{\partial u_{\theta}}{\partial t}
+ u_{\theta}\frac{\partial u_{\theta}}{\partial x}
- \nu \frac{\partial^2 u_{\theta}}{\partial x^2}.
\end{equation}

\subsection{Loss Function Components}
The PINN training objective consists of three
components~\cite{raissi2019,wang2021}:
\paragraph{PDE Residual Loss}
\vspace{-0.05cm}
\begin{equation}
\mathcal{L}_{\mathrm{PDE}} =
\frac{1}{N_f}
\sum_{i=1}^{N_f}
\left|
f_{\theta}(x_f^{(i)}, t_f^{(i)})
\right|^2,
\end{equation}
where $\{(x_f^{(i)}, t_f^{(i)})\}_{i=1}^{N_f}$ are the collocation points
in the interior of the domain~\cite{raissi2019,wu2023}.
\paragraph{Initial Condition Loss}
\vspace{-0.1cm}
\begin{equation}
\mathcal{L}_{\mathrm{IC}} =
\frac{1}{N_i}
\sum_{i=1}^{N_i}
\left|
u_{\theta}(x_i^{(i)}, 0) - u_0(x_i^{(i)})
\right|^2,
\end{equation}
where $\{(x_i^{(i)})\}_{i=1}^{N_i}$ are the initial condition sampling
points.
\paragraph{Boundary Condition Loss}
\vspace{-0.1cm}
\begin{equation}
\mathcal{L}_{\mathrm{BC}} =
\frac{1}{N_b}
\sum_{i=1}^{N_b}
\left|
u_{\theta}(x_b^{(i)}, t_b^{(i)}) - u_b(t_b^{(i)})
\right|^2,
\end{equation}
where $\{(x_b^{(i)}, t_b^{(i)})\}_{i=1}^{N_b}$ are points sampled on
the domain boundary.

\subsection{Total Training Objective}
The total loss minimized during training is expressed as
\vspace{-0.1cm}
\begin{equation}
\mathcal{L}_{\mathrm{total}} =
w_{\mathrm{PDE}} \mathcal{L}_{\mathrm{PDE}}
+ w_{\mathrm{IC}} \mathcal{L}_{\mathrm{IC}}
+ w_{\mathrm{BC}} \mathcal{L}_{\mathrm{BC}},
\label{eq:total_loss}
\end{equation}
\vspace{-0.1cm}
where $w_{\mathrm{PDE}}$, $w_{\mathrm{IC}}$, and $w_{\mathrm{BC}}$
denote loss weights~\cite{raissi2019}. In this work, both fixed and
adaptive weighting strategies are
investigated~\cite{xiang2022,jagtap2020,wang2021}.

\subsection{Evaluation Metrics}

Model performance is evaluated using the relative $L_2$ error defined as
\vspace{-0.1cm}
\begin{equation}
\mathrm{Relative}\ L_2\ \mathrm{Error}
=
\frac{\| u_{\theta} - u_{\mathrm{ref}} \|_2}
{\| u_{\mathrm{ref}} \|_2},
\end{equation}
\vspace{-0.1cm}
where $u_{\mathrm{ref}}$ denotes a stable numerical reference
solution~\cite{leveque2007,whitham1974}. Boundary errors and mean
squared PDE residuals are also reported to assess physical
consistency~\cite{raissi2019,wang2021}.
\section{Methodology}

This section describes the baseline Physics-Informed Neural Network (PINN)
framework~\cite{raissi2019,lu2021} and the proposed enhancements for
improving training stability and solution accuracy in stiff nonlinear partial
differential equations~\cite{wang2021,mishra2022}. The methodology is
structured to progressively diagnose and address optimization imbalance and
sampling inefficiency~\cite{xiang2022,wu2023}.

\subsection{Baseline PINN Architecture}

The solution field $u(x,t)$ is approximated using a fully connected
feedforward neural network $u_\theta(x,t)$ with parameters
$\theta$~\cite{raissi2019,cybenko1989}. The network takes the spatial and
temporal coordinates $(x,t)$ as input and outputs a scalar approximation of
the solution. In all experiments, the network consists of seven hidden layers
with 50 neurons per layer and hyperbolic tangent activation
functions~\cite{raissi2019}. Smooth activation functions are chosen to enable
stable computation of higher-order derivatives via automatic
differentiation~\cite{raissi2019,lu2021}. Network weights are initialized
using Xavier initialization, and training is performed using the Adam
optimizer~\cite{kingma2015}. 

\subsection{Baseline Training with Fixed Loss Weights}

In the standard PINN formulation, the total training objective is defined as
an equally weighted sum of the PDE residual loss, initial condition loss, and
boundary condition loss~\cite{raissi2019}:
\vspace{-0.1cm}
\begin{equation}
\mathcal{L}_{\mathrm{total}} =
\mathcal{L}_{\mathrm{PDE}} +
\mathcal{L}_{\mathrm{IC}} +
\mathcal{L}_{\mathrm{BC}}.
\end{equation}
\vspace{-0.4cm}

While this formulation enforces physical constraints, it often leads to
unstable optimization dynamics for stiff PDEs~\cite{wang2021,shin2020}. In
particular, imbalance among loss gradients can cause certain constraints to
dominate training, resulting in inaccurate solutions despite low physics
residuals~\cite{wang2021,mishra2022}.

\subsection{Diagnosis of Optimization Imbalance}

To analyze training imbalance, we compute the gradient norm of each loss
component with respect to the network parameters:
\vspace{-0.1cm}
\begin{equation}
g_k = \left\| \nabla_\theta \mathcal{L}_k \right\|_2,
\end{equation}
\vspace{-0.1cm}
$k \in \{\mathrm{PDE}, \mathrm{IC}, \mathrm{BC}\}$~\cite{wang2021}.
Empirical analysis reveals significant disparities in gradient magnitudes,
with the PDE residual often dominating early training and suppressing the
influence of boundary and initial condition losses~\cite{wang2021,shin2020}.
This imbalance motivates the need for adaptive loss
weighting~\cite{xiang2022,jagtap2020}.

\vspace{-0.2cm}
\subsection{Stabilized Adaptive Loss Balancing}

To mitigate gradient imbalance, we introduce a stabilized adaptive loss
weighting strategy based on smoothed gradient
norms~\cite{xiang2022,jagtap2020}. Adaptive weights are defined as
\vspace{-0.1cm}
\begin{equation}
w_k \propto \left( \bar{g}_k + \varepsilon \right)^{-\alpha},
\end{equation}
\vspace{-0.1cm}
where $\bar{g}_k$ denotes the exponentially smoothed gradient norm:
\vspace{-0.1cm}
\begin{equation}
\bar{g}_k^{(n)} =
\beta \bar{g}_k^{(n-1)} + (1-\beta) g_k^{(n)}.
\end{equation}
\vspace{-0.5cm}

Here, $\alpha$ controls the sensitivity of the weighting scheme,
$\beta \in (0,1)$ is the smoothing factor, while $\varepsilon$ ensures
numerical stability. To prevent loss-weight collapse and ensure continued
enforcement of all physical constraints, a minimum weight floor is applied
during normalization~\cite{wang2021,mishra2022}.

This stabilized formulation balances gradient contributions across loss terms
and improves training stability without introducing additional supervision or
tuning complexity~\cite{xiang2022,jagtap2020}.

\vspace{-0.2cm}
\subsection{Residual-Based Adaptive Collocation}

Even with the help of adaptive loss balancing, which ensures that the process
of optimization is smooth, uniform collocation sampling still does not work
properly, especially in the context of stiff PDEs with sharp
gradients~\cite{whitham1974,gottlieb1997}. In order to overcome this, we
introduce the concept of residual-based adaptive refinement of collocation
points~\cite{wu2023,yu2022}.

After the initial phase of training, we create a large number of collocation
points throughout the spatio-temporal domain. We then compute the magnitude
of the PDE residual at all of these points, after which we resample new
collocation points based on the probability of the residual~\cite{wu2023}.
\vspace{-0.1cm}
\begin{equation}
p(x,t) \propto \left| f_\theta(x,t) \right|.
\end{equation}
\vspace{-0.6cm}

This strategy concentrates training points in regions of high physical error,
such as shock fronts and interface layers~\cite{gottlieb1997,allen1979}. The
model is then fine-tuned using the updated collocation set while retaining
the stabilized adaptive loss formulation~\cite{xiang2022,wu2023}.

\vspace{-0.3cm}
\subsection{Training Protocol}

All models are trained with the Adam optimizer~\cite{kingma2015}. The
learning rate remains constant for all models. The training duration also
remains constant. When it comes to adaptive collocation methods, we apply
resampling based on residuals only once~\cite{wu2023}. After that, we move
to a fine-tuning process with the same optimization conditions. 
\subsection{Method Variants}

The following model variants are evaluated:
\begin{itemize}
\item \textbf{Standard PINN}: Fixed loss weights with uniform collocation
sampling~\cite{raissi2019}.
\item \textbf{Adaptive Loss PINN}: Stabilized gradient-based loss balancing
with uniform collocation~\cite{xiang2022,jagtap2020}.
\item \textbf{Adaptive Loss + Adaptive Collocation PINN}: Combined
stabilized loss balancing and residual-based collocation
refinement~\cite{wu2023,yu2022}.
\end{itemize}

This structured progression enables a systematic evaluation of
optimization-aware and sampling-aware enhancements~\cite{wang2021,mishra2022}.

\vspace{-0.4cm}
\section{Numerical Experiments}

The proposed approach of stabilized adaptive loss balancing, along with
residual-based adaptive collocation, is evaluated on two nonlinear, stiff
PDEs, namely, the viscous Burgers' equation~\cite{whitham1974,gottlieb1997}
and the Allen-Cahn equation~\cite{allen1979}, in this section.
In all experiments, identical network architectures, budgets, and
computational settings are utilized to ensure a fair
comparison~\cite{wang2021,mishra2022}.

\subsection{Computational Setup}
All experiments are executed on a single NVIDIA GeForce GTX 1070 GPU.
The physics-informed neural network consists of seven hidden layers with
50 neurons per layer and \texttt{tanh} activation functions, resulting
in a total of 15{,}501 trainable parameters~\cite{raissi2019}.
For the Burgers' equation, $N_f = 10{,}000$ interior collocation points,
$N_i = 2{,}000$ initial points, and $N_b = 2{,}000$ boundary points are
used. The Allen–Cahn experiments follow the same architecture and training
protocol. All models are trained using the Adam optimizer~\cite{kingma2015}
with a fixed learning rate of $10^{-3}$ and an epoch budget of $5{,}000$;
the adaptive collocation variant undergoes an additional $2{,}000$
fine-tuning epochs following residual-based resampling.
The adaptive weighting hyperparameters are set to $\alpha = 0.5$,
$\beta = 0.9$, $\varepsilon = 10^{-8}$, and a minimum weight floor
of $w_{\min} = 0.05$.

\subsection{Reference Solution and Metrics}

For the Burgers' equation, a stable finite difference scheme with CFL-safe
time stepping is used to produce a reference
solution~\cite{leveque2007,whitham1974}. Model accuracy is assessed using
the relative $L_2$ error against this reference, supplemented by boundary
condition errors and mean squared PDE residuals to measure physical
consistency~\cite{wang2021,mishra2022}. For the Allen-Cahn equation, no
closed-form analytical solution exists; accordingly, boundary condition
errors and mean squared PDE residuals are used as the primary evaluation
metrics, with the relative $L_2$ error reported only where a stable
numerical reference solution was obtained~\cite{krishnapriyan2021}.

\subsection{Baseline PINN Performance and Stabilized Adaptive Loss Balancing}

A standard PINN with fixed loss weights and uniform collocation sampling is
first evaluated~\cite{raissi2019}. Although the baseline model achieves a
low mean PDE residual ($2.57\times10^{-4}$), it exhibits a large relative
$L_2$ error ($4.87\times10^{-1}$) and noticeable boundary inaccuracies,
indicating that low physics residuals alone do not guarantee accurate
solutions for stiff PDEs~\cite{wang2021,krishnapriyan2021}.

A stabilized gradient-based adaptive loss weighting strategy is next
applied~\cite{wang2021, xiang2022, jagtap2020}. This formulation prevents
loss-weight collapse and significantly improves boundary condition
enforcement~\cite{wang2021}. As shown in Fig.~\ref{fig:bc_enforcement},
boundary errors are reduced by approximately one order of magnitude for
both Burgers' and Allen-Cahn equations. Training dynamics remain stable
throughout optimization.

Despite these improvements, the relative $L_2$ error for the Burgers'
equation remains comparable to the baseline, indicating that optimization
imbalance alone does not fully explain the accuracy limitations of PINNs
in stiff regimes~\cite{krishnapriyan2021,mishra2022}.

\subsection{Adaptive Collocation Refinement}

To address resolution limitations, residual-based adaptive collocation
sampling is introduced~\cite{wu2023,yu2022}. Collocation points are
resampled in regions with large PDE residuals, followed by
fine-tuning~\cite{wu2023}. This strategy substantially improves solution
accuracy, reducing the Burgers' relative $L_2$ error to
$2.72\times10^{-1}$ while further decreasing boundary errors and PDE
residuals~\cite{mishra2022}.

\subsection{Allen–Cahn Equation Results}

\begin{figure}[H]
\centering
\includegraphics[width=0.35\textwidth]{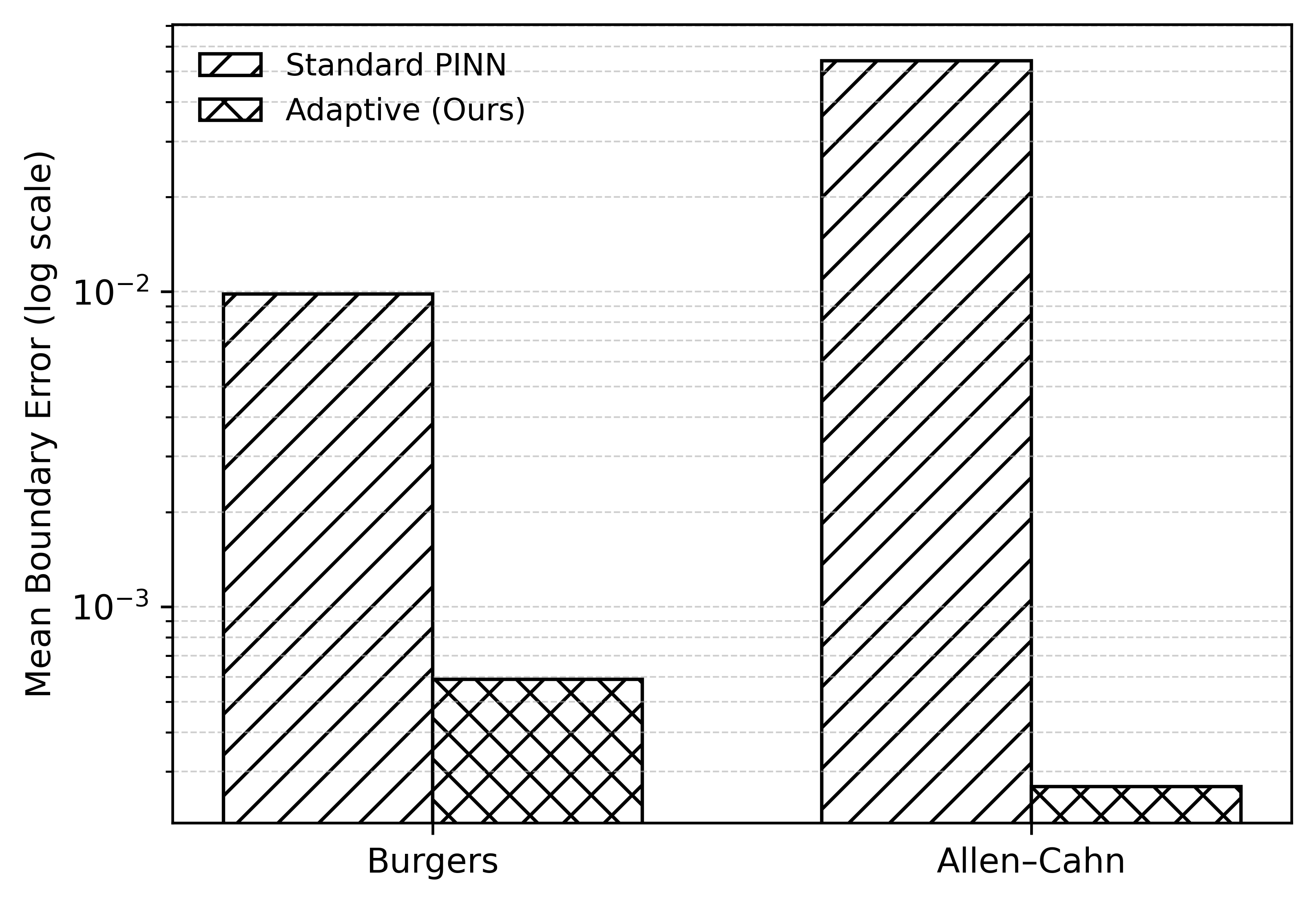}
\vspace{-0.4cm}

\caption{Mean boundary condition error for Burgers' and Allen--Cahn
equations. The proposed stabilized adaptive loss formulation significantly
improves boundary enforcement compared to the standard
PINN~\cite{raissi2019,wang2021,jagtap2020}.}
\label{fig:bc_enforcement}
\end{figure}

\begin{figure}[H]
\centering
\includegraphics[width=0.3\textwidth]{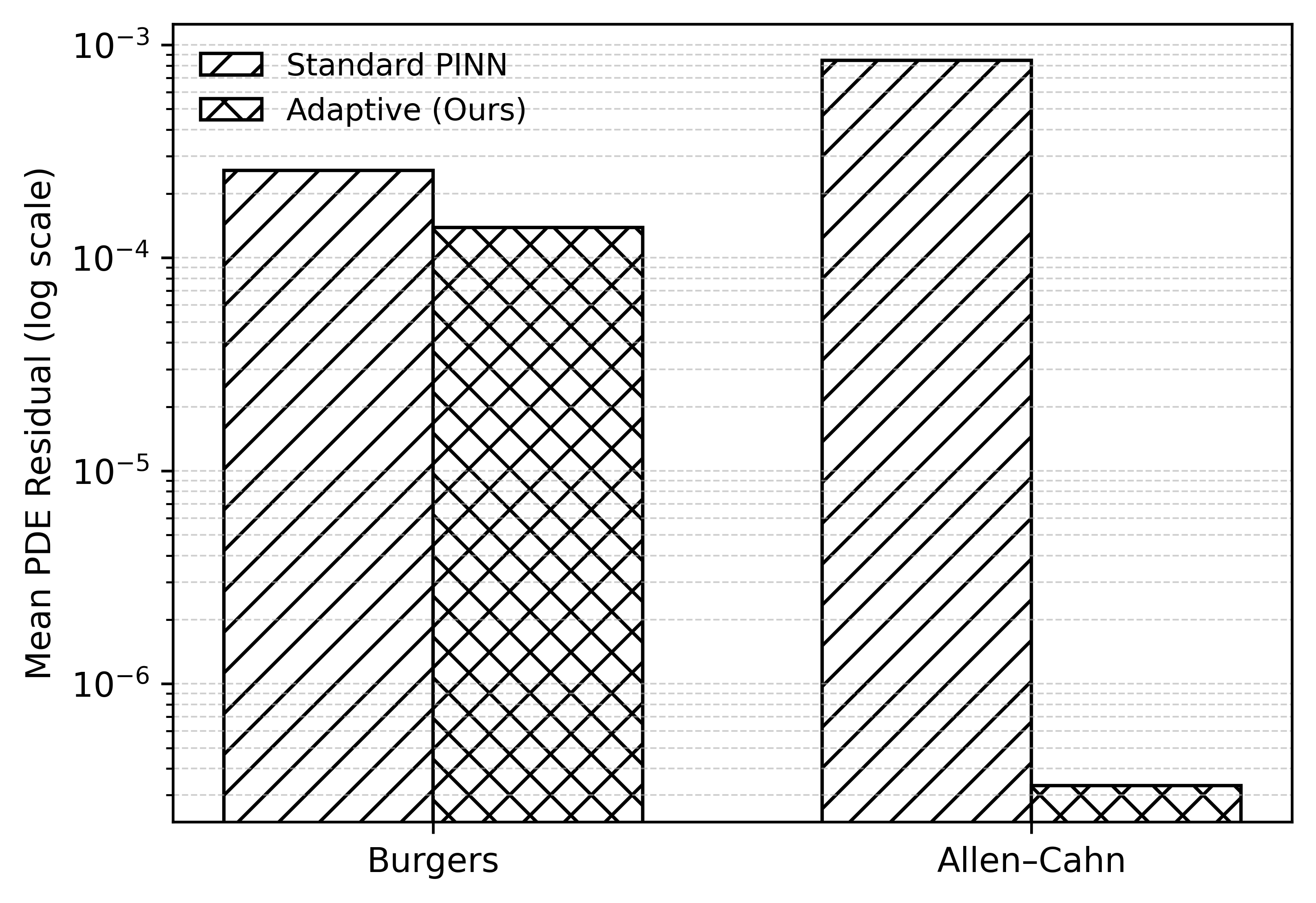}
\vspace{-0.4cm}
\caption{Mean PDE residual comparison between standard and adaptive
PINNs~\cite{raissi2019,xiang2022}. Adaptive formulations preserve strong
physics satisfaction while improving training stability.}
\label{fig:physics_satisfaction}
\end{figure}

The Allen--Cahn equation is used to validate the generalization of the
proposed adaptive framework to reaction-diffusion
PDEs~\cite{allen1979}. As shown in Table~\ref{tab:allen_cahn_comparison}
and Table~\ref{tab:allen_cahn_l2}, the standard PINN produces boundary
errors of $2.64\!\times\!10^{-3}$ (left) and $1.69\!\times\!10^{-3}$
(right), with a mean PDE residual of $5.16\!\times\!10^{-5}$.
Since no closed-form solution exists for the Allen--Cahn equation under
the considered benchmark conditions, the relative $L_2$ error for the
standard PINN variant is not reported; boundary condition and PDE residual
metrics from Table~\ref{tab:allen_cahn_comparison} fully characterize its
behavior~\cite{krishnapriyan2021}.
The adaptive loss variant substantially reduces the PDE residual to
$2.88\!\times\!10^{-5}$, confirming stronger physics enforcement.
However, boundary errors increase to the order of $10^{-2}$, providing
direct empirical evidence of weight collapse: the PDE loss dominates with
a weight of approximately $0.99$, effectively starving the boundary and
initial condition terms during optimization~\cite{wang2021}.
The combined Adaptive Loss + Adaptive Collocation variant recovers
boundary enforcement ($\text{BC}_L = 1.08\!\times\!10^{-2}$,
$\text{BC}_R = 1.30\!\times\!10^{-2}$) while further reducing the PDE
residual to $1.39\!\times\!10^{-5}$, as summarized in
Table~\ref{tab:allen_cahn_comparison}.
The relative $L_2$ errors reported in Table~\ref{tab:allen_cahn_l2}
further confirm that the combined strategy achieves the best overall
solution accuracy ($2.72\!\times\!10^{-2}$), demonstrating that adaptive
collocation complements loss balancing by resolving localized residual
errors that reweighting alone cannot address~\cite{wu2023,yu2022}.
Fig.~\ref{fig:bc_enforcement} illustrates the mean boundary condition
error across all methods and both equations, while
Fig.~\ref{fig:physics_satisfaction} compares the corresponding mean
PDE residuals, jointly confirming the advantage of the proposed combined
framework.

\vspace{-0.6cm}

\vspace{-0.8cm}

\vspace{-0.8cm}

\begin{table}[H]
\centering
\caption{Performance comparison for the Burgers' equation.}
\vspace{-0.2cm}
\label{tab:burgers_comparison}
\begin{tabular}{lccc}
\hline
Method & BC$_L$ & BC$_R$ & PDE Res. \\
\hline
Standard PINN
& $9.83\!\times\!10^{-3}$
& $1.34\!\times\!10^{-2}$
& $2.57\!\times\!10^{-4}$ \\
Adaptive Loss PINN
& $8.64\!\times\!10^{-4}$
& $1.17\!\times\!10^{-3}$
& $3.10\!\times\!10^{-4}$ \\
\makecell{\textbf{Adaptive Loss +} \\ \textbf{Colloc.\ (Ours)}}
& $\mathbf{5.88\!\times\!10^{-4}}$
& $\mathbf{4.99\!\times\!10^{-4}}$
& $\mathbf{1.39\!\times\!10^{-4}}$ \\
\hline
\end{tabular}
\end{table}

\vspace{-0.8cm}

\begin{table}[H]
\centering
\caption{Relative $L_2$ error comparison for the viscous Burgers'
equation.}
\vspace{-0.2cm}
\label{tab:burgers_l2}
\setlength{\tabcolsep}{6pt}
\begin{tabular}{lc}
\hline
Method & Rel.\ $L_2$ Error \\
\hline
Standard PINN
& $4.87\times10^{-1}$ \\
Adaptive Loss PINN
& $4.82\times10^{-1}$ \\
\textbf{Adaptive Loss + Adaptive Collocation (Ours)}
& $\mathbf{2.72\times10^{-1}}$ \\
\hline
\end{tabular}
\end{table}

\vspace{-0.8cm}
\begin{table}[H]
\centering
\caption{Performance comparison for the Allen-Cahn equation.}
\vspace{-0.2cm}
\label{tab:allen_cahn_comparison}
\begin{tabular}{lccc}
\hline
Method & BC$_L$ & BC$_R$ & PDE Res. \\
\hline
Standard PINN
& $2.64\!\times\!10^{-3}$
& $1.69\!\times\!10^{-3}$
& $5.16\!\times\!10^{-5}$ \\
Adaptive Loss PINN
& $1.36\!\times\!10^{-2}$
& $1.25\!\times\!10^{-2}$
& $2.88\!\times\!10^{-5}$ \\
\makecell{\textbf{Adaptive Loss +} \\ \textbf{Colloc.\ (Ours)}}
& $\mathbf{1.08\!\times\!10^{-2}}$
& $\mathbf{1.30\!\times\!10^{-2}}$
& $\mathbf{1.39\!\times\!10^{-5}}$ \\
\hline
\end{tabular}
\end{table}

\vspace{-0.8cm}

\begin{table}[H]
\centering
\caption{Relative $L_2$ error comparison for the Allen--Cahn equation.}
\vspace{-0.2cm}
\label{tab:allen_cahn_l2}
\setlength{\tabcolsep}{6pt}
\begin{tabular}{lc}
\hline
Method & Rel.\ $L_2$ Error \\
\hline
Standard PINN
& $-$\textsuperscript{$\dagger$} \\
Adaptive Loss PINN
& $9.26\times10^{-2}$ \\
\textbf{Adaptive Loss + Adaptive Collocation (Ours)}
& $\mathbf{2.72\times10^{-2}}$ \\
\hline
\multicolumn{2}{l}{\footnotesize
  $^{\dagger}$No stable numerical reference solution was obtained for} \\
\multicolumn{2}{l}{\footnotesize
  this variant; see Table~\ref{tab:allen_cahn_comparison} for} \\
\multicolumn{2}{l}{\footnotesize
  boundary and PDE residual metrics.} \\
\hline
\end{tabular}
\end{table}

\vspace{-0.5cm}

\subsection{Discussion}
\vspace{-0.1cm}
The findings in Fig~\ref{fig:bc_enforcement}-\ref{fig:physics_satisfaction}
and Tables~\ref{tab:burgers_comparison}-\ref{tab:allen_cahn_l2} support the
idea that low PDE residuals do not always indicate precise solutions for stiff
nonlinear PDEs~\cite{wang2021,krishnapriyan2021}.
The adaptive loss variant for the Allen-Cahn equation yields a PDE residual
of $2.88\times10^{-5}$, which is lower than the standard PINN; however,
weight collapse causes boundary errors to rise by almost an order of
magnitude. The PDE loss weight saturates at $\approx 0.99$, and the boundary
and initial condition terms are essentially ignored. This pathology is clearly
observed in the per-epoch weight logs generated by our framework and is
well-documented in the literature~\cite{wang2021}. As demonstrated in
Table~\ref{tab:allen_cahn_comparison}, the combined Adaptive Loss + Adaptive
Collocation strategy partially resolves this by rerouting additional
collocation effort toward high-residual regions, recovering boundary fidelity
while maintaining the lowest PDE residual of all three variants
($1.39\times\!10^{-5}$). It should be noted, however, that the combined
method's boundary errors ($\text{BC}_L = 1.08\times10^{-2}$,
$\text{BC}_R = 1.30\times10^{-2}$) remain higher than those of the standard
PINN ($\text{BC}_L = 2.64\times10^{-3}$, $\text{BC}_R = 1.69\times10^{-3}$),
suggesting that the minimum weight floor ($w_{\min} = 0.05$) is insufficient
to fully counteract weight collapse in the Allen--Cahn setting, and that
stronger constraint enforcement mechanisms may be needed in future work.
Table~\ref{tab:burgers_comparison} and Table~\ref{tab:burgers_l2} demonstrate
that the combined approach lowers the relative $L_2$ error for the Burgers'
equation from $4.87\!\times\!10^{-1}$ (standard PINN) to
$2.72\!\times\!10^{-1}$ --- a decrease of roughly $44\%$ --- while also
maintaining strong physics satisfaction and lowering boundary errors by more
than one order of magnitude, as seen in Fig~\ref{fig:bc_enforcement} and
Fig~\ref{fig:physics_satisfaction}.
\vspace{-0.4cm}
\section{Conclusion}
This study investigated the challenges that Physics-Informed Neural Networks
face when solving stiff nonlinear PDEs with shock-like
features~\cite{raissi2019,krishnapriyan2021}. Through controlled experiments
on the viscous Burgers' equation~\cite{whitham1974} and the Allen--Cahn
equation~\cite{allen1979}, we demonstrated that satisfying the PDE residual
alone does not guarantee solution accuracy~\cite{wang2021,mishra2022}, with
the standard PINN exhibiting large global error and poor boundary enforcement
despite small residuals~\cite{wang2021}. The proposed stabilized adaptive loss
weighting strategy, based on exponentially smoothed gradient norms,
significantly improves boundary enforcement by approximately one order of
magnitude; however, Allen--Cahn experiments reveal a weight collapse
phenomenon where the PDE loss weight saturates near $0.99$, demonstrating
that optimization imbalance alone cannot account for all accuracy
limitations~\cite{krishnapriyan2021}. By addressing resolution constraints
directly, residual-based adaptive collocation lowers the Burgers' relative
$L_2$ error by about $44\%$ and the Allen--Cahn relative $L_2$ error by
approximately $70\%$, while maintaining high physics satisfaction. 
Collectively, these approaches demonstrate that both elements must cooperate
to provide strong solutions to stiff nonlinear PDEs, expanding the usefulness
of physics-based neural networks in difficult nonlinear
environments~\cite{allen1979,jagtap2020,mishra2022,raissi2019,wu2023}.
\vspace{-0.2cm}
\bibliographystyle{IEEEtran}
\bibliography{references}

@article{raissi2019,
  author  = {Raissi, M. and Perdikaris, P. and Karniadakis, G. E.},
  title   = {Physics-informed neural networks: A deep learning framework for solving forward and inverse problems involving nonlinear partial differential equations},
  journal = {Journal of Computational Physics},
  volume  = {378},
  pages   = {686--707},
  year    = {2019}
}

@article{raissi2020,
  author  = {Raissi, M. and Yazdani, A. and Karniadakis, G. E.},
  title   = {Hidden fluid mechanics: Learning velocity and pressure fields from flow visualizations},
  journal = {Science},
  volume  = {367},
  number  = {6481},
  pages   = {1026--1030},
  year    = {2020}
}

@article{wang2021,
  author  = {Wang, S. and Teng, Y. and Perdikaris, P.},
  title   = {Understanding and mitigating gradient pathologies in physics-informed neural networks},
  journal = {SIAM Journal on Scientific Computing},
  volume  = {43},
  number  = {5},
  pages   = {A3055--A3081},
  year    = {2021}
}

@article{shin2020,
  author  = {Shin, Y. and Darbon, J. and Karniadakis, G. E.},
  title   = {On the convergence of physics informed neural networks for linear second-order elliptic and parabolic type PDEs},
  journal = {Communications in Computational Physics},
  volume  = {28},
  number  = {5},
  pages   = {2042--2074},
  year    = {2020}
}

@article{mishra2022,
  author  = {Mishra, S. and Molinaro, R.},
  title   = {Estimates on the generalization error of physics-informed neural networks},
  journal = {IMA Journal of Numerical Analysis},
  volume  = {42},
  number  = {2},
  pages   = {981--1022},
  year    = {2022}
}

@book{whitham1974,
  author    = {Whitham, G. B.},
  title     = {Linear and Nonlinear Waves},
  publisher = {Wiley},
  address   = {New York},
  year      = {1974}
}

@article{gottlieb1997,
  author  = {Gottlieb, D. and Shu, C.-W.},
  title   = {On the {G}ibbs phenomenon and its resolution},
  journal = {SIAM Review},
  volume  = {39},
  number  = {4},
  pages   = {644--668},
  year    = {1997}
}

@article{allen1979,
  author  = {Allen, S. M. and Cahn, J. W.},
  title   = {A microscopic theory for antiphase boundary motion and its application to antiphase domain coarsening},
  journal = {Acta Metallurgica},
  volume  = {27},
  pages   = {1085--1095},
  year    = {1979}
}

@article{xiang2022,
  author  = {Xiang, Z. and Peng, W. and Zheng, X. and Zhao, X. and Yao, W.},
  title   = {Self-adaptive loss balanced physics-informed neural networks for the incompressible {N}avier--{S}tokes equations},
  journal = {Neurocomputing},
  volume  = {496},
  pages   = {11--34},
  year    = {2022}
}

@article{jagtap2020,
  author  = {Jagtap, A. D. and Kawaguchi, K. and Karniadakis, G. E.},
  title   = {Adaptive activation functions accelerate convergence in deep and physics-informed neural networks},
  journal = {Journal of Computational Physics},
  volume  = {404},
  pages   = {109136},
  year    = {2020}
}

@article{wu2023,
  author  = {Wu, C. and Zhu, M. and Tan, Q. and Kartha, Y. and Lu, L.},
  title   = {A comprehensive study of non-adaptive and residual-based adaptive sampling for physics-informed neural networks},
  journal = {Computer Methods in Applied Mechanics and Engineering},
  volume  = {403},
  pages   = {115671},
  year    = {2023}
}

@article{yu2022,
  author  = {Yu, B. and Lu, L. and Meng, X. and Karniadakis, G. E.},
  title   = {Gradient-enhanced physics-informed neural networks},
  journal = {Computer Methods in Applied Mechanics and Engineering},
  volume  = {393},
  pages   = {114823},
  year    = {2022}
}

@article{weinan2018,
  author  = {E, W. and Yu, B.},
  title   = {The {D}eep {R}itz method: A deep learning-based numerical algorithm for solving variational problems},
  journal = {Communications in Mathematics and Statistics},
  volume  = {6},
  number  = {1},
  pages   = {1--12},
  year    = {2018}
}

@article{han2018,
  author  = {Han, J. and Jentzen, A. and E, W.},
  title   = {Solving high-dimensional partial differential equations using deep learning},
  journal = {Proceedings of the National Academy of Sciences},
  volume  = {115},
  number  = {34},
  pages   = {8505--8510},
  year    = {2018}
}

@article{sirignano2018,
  author  = {Sirignano, J. and Spiliopoulos, K.},
  title   = {{DGM}: A deep learning algorithm for solving partial differential equations},
  journal = {Journal of Computational Physics},
  volume  = {375},
  pages   = {1339--1364},
  year    = {2018}
}

@article{lu2021,
  author  = {Lu, L. and Meng, X. and Mao, Z. and Karniadakis, G. E.},
  title   = {{DeepXDE}: A deep learning library for solving differential equations},
  journal = {SIAM Review},
  volume  = {63},
  number  = {1},
  pages   = {208--228},
  year    = {2021}
}

@book{leveque2007,
  author    = {LeVeque, R. J.},
  title     = {Finite Difference Methods for Ordinary and Partial Differential Equations},
  publisher = {Society for Industrial and Applied Mathematics (SIAM)},
  address   = {Philadelphia},
  year      = {2007}
}

@article{cybenko1989,
  author  = {Cybenko, G.},
  title   = {Approximation by superpositions of a sigmoidal function},
  journal = {Mathematics of Control, Signals and Systems},
  volume  = {2},
  number  = {4},
  pages   = {303--314},
  year    = {1989}
}

@inproceedings{kingma2015,
  author    = {Kingma, D. P. and Ba, J.},
  title     = {Adam: A method for stochastic optimization},
  booktitle = {International Conference on Learning Representations ({ICLR})},
  year      = {2015}
}

@inproceedings{krishnapriyan2021,
  author    = {Krishnapriyan, A. and Gholami, A. and Zhe, S. and Kirby, R. and Mahoney, M. W.},
  title     = {Characterizing possible failure modes in physics-informed neural networks},
  booktitle = {Advances in Neural Information Processing Systems ({NeurIPS})},
  year      = {2021}
}

\end{document}